\documentclass{svproc}
\usepackage{url}

\usepackage{graphicx}
\usepackage{float}
\usepackage[colorlinks=true, linkcolor=blue, citecolor=blue, urlcolor=blue]{hyperref}

\usepackage{amsmath}
\usepackage{graphicx}

\usepackage{textcomp}
\usepackage{soul}
\usepackage{xcolor}
\usepackage[most]{tcolorbox}
\usepackage{tabularx}
\usepackage{booktabs}   
\usepackage{array}
\usepackage{float}      
\usepackage{placeins}   

\newcolumntype{L}{>{\raggedright\arraybackslash}X}

\def\BibTeX{{\rm B\kern-.05em{\sc i\kern-.025em b}\kern-.08em
    T\kern-.1667em\lower.7ex\hbox{E}\kern-.125emX}}

\begin{document}
\mainmatter              
\title{Usable Is Not Enough: Trust in Unchecked AI Synthesis in CoSLR, a Human--AI System for Systematic Literature Reviews}
\titlerunning{Trust in Unchecked AI Synthesis in CoSLR}  

\author{
Md Aidul Islam \and
Malik Abdul Sami \and
Muhammad Waseem \and
Zeeshan Rasheed \and
Kai-Kristian Kemell \and
Zheying Zhang \and
Pekka Abrahamsson
}

\authorrunning{M. A. Islam et al.}

\institute{
Faculty of Information Technology and Communication Sciences,\\
Tampere University, Tampere, Finland\\
\email{mdaidul.islam@tuni.fi, malik.sami@tuni.fi}\\
\email{muhammad.waseem@tuni.fi, zeeshan.rasheed@tuni.fi}\\
\email{kai-kristian.kemell@tuni.fi, zheying.zhang@tuni.fi}\\
\email{pekka.abrahamsson@tuni.fi}
}

\maketitle              

\begin{abstract}
Systematic literature reviews (SLRs) are central to evidence-based research but remain labour-intensive across planning, screening, analysis, and reporting. Large language models (LLMs) can now produce fluent, citation-bearing review text, making it hard to tell verified synthesis from synthesis that only appears authoritative. Human--AI systems therefore rely on researchers checking generated output before accepting it. We present CoSLR, a multi-agent system that supports the SLR workflow through a three-phase pipeline combining LLM agents with retrieval-augmented generation (RAG), and places mandatory human checkpoints between generated output and its acceptance. In an exploratory survey study, 63 participants used CoSLR for about 15 minutes. They rated it positively for usability (grouped mean 3.78 on a five-point scale), especially for research-question formulation, responsiveness, and report generation, and most expected it to save time. However, trust in AI-generated summaries without additional human checking was the lowest-rated of the positively worded items (mean 3.02), and 35\% of participants rated it 4 or 5, a rating positively associated with expected time savings (Spearman $\rho = 0.30$). Open-ended feedback called for clearer evidence traceability and stronger manual checkpoints. These findings suggest that a usable interface with mandatory checkpoints does not by itself ensure that oversight takes place, and that checkpoint design---how much verification a checkpoint demands---should be treated as an explicit design variable for Human--AI research tools.
\end{abstract}
\section{Introduction}
\label{sec:introduction}

SLRs are widely used to synthesize existing research in a structured, rigorous, and reproducible way \cite{kitchenham2007guidelines}. They help researchers identify what is already known, compare evidence across studies, and locate gaps for future research. In software engineering and related fields, SLRs are especially important because research evidence is often distributed across many venues, methods, and application domains \cite{kitchenham2007guidelines}, and their value depends on transparent, well-reported synthesis so that results can be assessed and reproduced \cite{page2021prisma}. However, conducting an SLR remains difficult in practice. Researchers must define review objectives, design search strategies, screen large sets of publications, extract relevant findings, and synthesize evidence into a coherent report \cite{kitchenham2007guidelines}. As the volume of scientific publications continues to grow, this process becomes increasingly time-consuming, labor-intensive, and vulnerable to inconsistency and human bias.

Recent advances in Artificial Intelligence (AI), especially LLMs, create new opportunities to assist researchers in literature review workflows. LLMs have shown general capabilities on tasks such as summarization, question answering, and information extraction \cite{bommasani2021opportunities,brown2020language}, and recent work has applied them specifically to title--abstract screening and screening decision support in systematic reviews \cite{wang2024zero,huotala2024promise}. Recent work has also explored agent-based systems for SLR automation, where multiple LLM-based agents support tasks such as search formulation, screening, extraction, and synthesis \cite{sami2024system,rouzrokh2025lattereview}. In parallel, Retrieval-Augmented Generation (RAG) has been used to ground language model outputs in retrieved evidence, which is important for reducing unsupported claims in literature-based workflows \cite{lewis2020retrieval}. These developments show clear potential for assisting several SLR stages. At the same time, SLR decisions still require human judgment, especially when defining the review scope, assessing relevance, validating extracted evidence, and interpreting findings. This motivates the need for Human--AI collaborative systems that assist researchers without replacing methodological control.

The authority of a systematic review rests on its method rather than its appearance: a documented search, explicit inclusion criteria, and a researcher
who has read the evidence and taken responsibility for the resulting claims. Large language models unsettle this because they can now produce review text that is fluent, well structured, and citation-bearing, so that a synthesis which was never verified can be difficult to distinguish from one that was. The concern is therefore not only that a model may occasionally err, but that unverified AI-generated synthesis may enter the scholarly record carrying the credibility of a systematic review, where subsequent reviews, researchers, and models can build on it and propagate the error. In a Human--AI workflow, the safeguard against this outcome is the researcher who is expected to check generated output before accepting it. This makes \emph{trust calibration}---the alignment between how much a researcher trusts AI-generated output and how much that output actually warrants verification---central to such systems \cite{okamura2020adaptive,ueno2022trust}. A mandatory checkpoint enforces the opportunity to verify, but it functions as a safeguard only if researchers use it to verify rather than to confirm; a checkpoint that is accepted without scrutiny provides the appearance of oversight without its substance \cite{okamura2020adaptive}. Whether interfaces of this kind produce genuine verification or only its appearance is therefore an open question, and this paper takes a first, exploratory step toward it.

However, the main challenge is not only automating individual tasks, but supporting the SLR process as a connected workflow. Existing tools and approaches have shown value in activities such as screening, prioritization, or document management, but researchers still need to move across multiple tools and manually connect outputs from one stage to the next \cite{khalil2022tools,van2021automation}. This creates a practical gap between task-level automation and a usable end-to-end Human--AI collaborative process for SLRs. In particular, there is still limited support for systems that combine automation, evidence grounding, and human oversight across planning, retrieval, analysis, and reporting.

This paper addresses that gap by presenting a Human--AI collaborative multi-agent system for assisting stages of the SLR workflow. The proposed system integrates a modular, multi-phase architecture that supports research objective formulation, search string generation, literature retrieval, document analysis, and structured report generation. To improve reliability, the system incorporates Retrieval-Augmented Generation (RAG) to ground generated outputs in retrieved document evidence and mitigate the risk of unsupported claims~ \cite{lewis2020retrieval}. The system also follows a human-in-the-loop design, allowing users to review, refine, and validate AI-generated outputs throughout the workflow.

To evaluate CoSLR, we conducted, a survey-based user study with 63 participants, including postgraduate students, researchers, and industry professionals. The evaluation combined quantitative Likert-scale responses and qualitative open-ended feedback. The results indicate positive user perceptions of usefulness, usability, and perceived efficiency,   while about a third of participants reported high trust in generated output without additional checking.

The main contributions of this paper are as follows:
\begin{itemize}
\item CoSLR, a Human--AI collaborative multi-agent system that integrates research-question formulation, search-string generation, retrieval, document analysis, RAG-based question answering, and report generation in a single workflow, with mandatory human checkpoints at the key review decisions.
\item An exploratory user study ($N=63$) of the system's perceived usefulness, usability, and expected efficiency, including items that allowed neutral or negative responses.
\item An analysis of participants' self-reported trust in AI-generated synthesis without additional checking, showing that about a third reported high trust despite mandatory checkpoints, and design implications for checkpoints and evidence provenance in Human--AI research tools.
\item An open replication package containing the anonymised survey data, questionnaire, system configuration, and analysis notebook.
\end{itemize}
\section{Background \& Related Work}
\label{sec:related_work}

Research on SLR automation has progressed from classical machine learning for screening to LLM-based support for multiple review stages. Reviews by Khalil et al.~\cite{khalil2022tools} and van Dinter et al.~\cite{van2021automation} show that automation can reduce manual effort, but that tools remain fragmented and typically support a single stage, such as screening, prioritization, or document management. Table~\ref{tab:relatedwork} summarizes representative tools and their main limitations.

\begin{table}[htbp]
\centering
\caption{Comparison of representative works on systematic literature review automation.}
\label{tab:relatedwork}
\footnotesize
\renewcommand{\arraystretch}{1.15}
\begin{tabularx}{\textwidth}{@{}
  >{\raggedright\arraybackslash\hsize=0.8\hsize}X
  >{\raggedright\arraybackslash\hsize=0.9\hsize}X
  >{\raggedright\arraybackslash\hsize=1.0\hsize}X
  >{\raggedright\arraybackslash\hsize=1.3\hsize}X @{}}
\toprule
\textbf{Study / Tool} & \textbf{Automation Focus} & \textbf{Techniques Used} & \textbf{Main Limitation} \\
\midrule
Abstrackr~\cite{wallace2012abstrackr} & Title and abstract screening & Supervised machine learning with user feedback & Requires substantial manual labeling and offers limited explainability. \\
\addlinespace
ASReview~\cite{van2020asreview} & Screening prioritization & Active learning with text-based ranking & Automates mainly one phase of the review workflow. \\
\addlinespace
Rayyan~\cite{ouzzani2016rayyan} & Collaborative screening support & Workflow management and relevance filtering & Primarily a decision-support platform rather than full automation. \\
\addlinespace
DistillerSR~\cite{hamel2020evaluation} & Review workflow management & Machine learning-assisted prioritization & Depends on human review for final decisions. \\
\addlinespace
RobotReviewer ~\cite{marshall2016robotreviewer} & Risk-of-bias assessment and extraction & NLP and machine learning & Strongly domain-specific to biomedical evidence synthesis. \\
\addlinespace
EPPI-Reviewer~\cite{eppireviewer2025} & Screening and review management & Automation-supported review workflow tools & Limited end-to-end intelligence across the full review lifecycle. \\
\addlinespace
van Dinter et al.~\cite{van2021automation} & Survey of automation methods & NLP and text mining techniques & Largely descriptive; does not propose a unified framework. \\
\addlinespace
Khalil et al.~\cite{khalil2022tools} & Scoping review of tools & Comparative review of automation software & Shows fragmentation of tools and limited pipeline integration. \\
\addlinespace
LatteReview~\cite{rouzrokh2025lattereview} & Multi-agent SLR screening and assessment & LLM-based reviewer agents with configurable roles & Early-stage framework; limited evaluation of human oversight in the workflow. \\
\bottomrule
\end{tabularx}
\end{table}




\subsection{Machine Learning Tools for SLR Support}

Abstrackr~\cite{wallace2012abstrackr} pioneered machine-learning prioritization of records based on reviewer feedback~\cite{wallace2010semi}, showing that screening workload can be reduced while maintaining acceptable recall, and ASReview~\cite{van2020asreview} extended this through active learning. Rayyan~\cite{ouzzani2016rayyan} and DistillerSR~\cite{hamel2020evaluation} support collaborative screening and workflow management, and RobotReviewer~\cite{marshall2016robotreviewer} automates risk-of-bias assessment for clinical trials. These tools reduce effort within individual stages, but search strategy development, synthesis, and interpretation remain largely manual, and several are tied to biomedical evidence synthesis.




\subsection{LLM- and Agent-Based Support}

LLMs perform well on summarization, question answering, and information extraction~\cite{bommasani2021opportunities,brown2020language}, and recent studies have applied them to title--abstract screening with promising results~\cite{huotala2024promise,wang2024zero}. Joos et al.~\cite{joos2026leveraging} use multiple LLMs for semi-automatic corpus filtration in SLRs, combining model-based classification with human supervision through an interactive visual-analytics interface; Ye et al.~\cite{ye2024hybrid} propose a hybrid semi-automated review workflow in which LLMs support title/abstract screening, full-text screening, and data extraction while researchers retain responsibility for inclusion/exclusion decisions and quality control. Both approaches maintain human involvement in key decisions, but target different portions of the review pipeline. Multi-agent frameworks such as LatteReview~\cite{rouzrokh2025lattereview} coordinate several LLM agents for screening and assessment, and frameworks such as PaperOrchestra~\cite{song2026paperorchestra} extend agents to broader research tasks including writing and synthesis. Retrieval-augmented generation (RAG) grounds model output in retrieved documents~\cite{lewis2020retrieval}, although its performance varies considerably across domains~\cite{han2024rag}.

Commercial research assistants such as Elicit\footnote{\url{https://elicit.com}}, Consensus\footnote{\url{https://consensus.app}}, and Perplexity\footnote{\url{https://www.perplexity.ai}}, as well as general-purpose LLMs with web search, already combine retrieval with cited answers. They are optimized for fast question answering, however, rather than for the protocol-driven SLR process, in which objectives, search strings, and inclusion decisions must be documented and approved by the researcher at defined points.

\subsection{Human Oversight of AI Output}

Because SLR screening is recall-sensitive, full automation without researcher validation remains risky~\cite{kitchenham2007guidelines,page2021prisma}. Work on human--AI interaction shows that oversight must be designed rather than assumed. Amershi et al.~\cite{amershi2019guidelines} recommend making system limitations explicit and supporting efficient correction; Bu\c{c}inca et al.~\cite{buccinca2021trust} show that cognitive forcing interventions reduce over-reliance on AI but are rated less favourably than low-effort interfaces; and Bansal et al.~\cite{bansal2019updates} show that higher AI accuracy does not automatically improve human--AI team performance. Appropriate reliance depends on aligning users' trust with the actual reliability of the system~\cite{okamura2020adaptive,ueno2022trust}.

\subsection{Research Gap}
\label{subsec:research_gap}

Existing SLR tools are fragmented across stages~\cite{khalil2022tools,van2021automation}; LLM-based approaches raise concerns about hallucination, prompt sensitivity, and reproducibility~\cite{wang2024zero,huotala2024promise,barnett2024seven}; and multi-agent SLR systems remain early-stage and only partially evaluated~\cite{sami2024system,rouzrokh2025lattereview}. Commercial assistants provide cited answers but not a protocol-driven workflow with explicit researcher approval. Moreover, systems that include human validation are, to our knowledge, rarely evaluated for whether users report trusting AI output without checking it. This paper addresses these gaps with a multi-agent system that integrates multiple SLR stages around mandatory human checkpoints, and with a user study of both its usability and participants' trust in unchecked output.
\section{Proposed System}
\label{sec:proposed_system}

\subsection{System Overview}
\begin{figure}[htbp]
\centering
\includegraphics[width=0.85\linewidth]{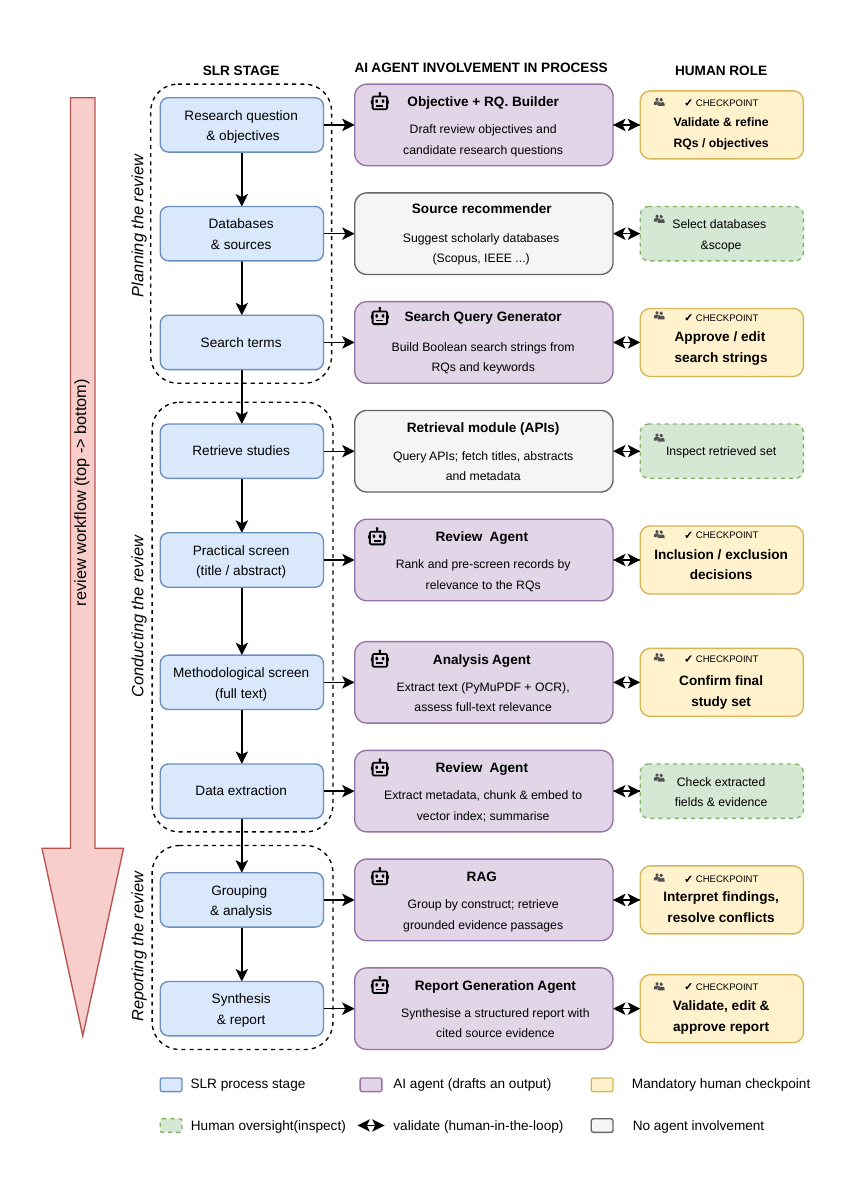}
\caption{The SLR process and the points of Human--AI collaboration in CoSLR, following the Kitchenham and Charters lifecycle~\cite{kitchenham2007guidelines}.}
\label{fig:slr_process}
\end{figure}

CoSLR follows the SLR workflow of Kitchenham and Charters, organized around planning, conducting, and reporting~\cite{kitchenham2007guidelines}. Figure~\ref{fig:slr_process} maps the Human--AI collaboration onto this lifecycle: for each stage, the centre column names the AI agent that drafts an output and the right column states the researcher's role. Amber boxes mark mandatory checkpoints, where a researcher decision is required before the workflow proceeds; dashed boxes mark lighter oversight. Three principles guide the design: transparency, so that automated actions produce inspectable outputs; human-in-the-loop control, so that researchers validate or modify generated content; and evidence grounding, so that outputs remain anchored in retrieved sources~\cite{lewis2020retrieval}.

The architecture (Fig.~\ref{fig:architecture} in the appendix) has three layers. A web-based user interaction layer lets researchers define objectives, review and upload papers, and interact with generated outputs. A processing layer invokes task-specific LLM agents for objective and research-question generation, search-string construction, document review and analysis, and report generation. All generative agents are implemented on OpenAI's GPT-4o; auxiliary preprocessing helpers call GPT-3.5-turbo, with prompts and model settings fixed for the study and provided in the replication package. The agents are called by the application logic rather than communicating directly, so each can be updated or replaced independently. A data management layer stores retrieved papers, metadata, embeddings, and intermediate artifacts, and logs research questions, queries, and retrieved chunks for traceability.

\subsection{Three-Phase Workflow}

In every phase, shown as the three columns of Fig.~\ref{fig:architecture} in the appendix, an agent produces a draft artifact, the researcher reviews and edits it, and the workflow advances only once the researcher confirms the result.

\textbf{Phase 1: Planning and search.} From an initial research prompt, the \emph{Objective Builder} identifies the topic, scope, and key concepts and drafts candidate objectives; the \emph{Research Questions Builder} derives aligned research questions; and the \emph{Search Query Generator} expands key concepts into synonym groups and builds database-compatible Boolean strings, optionally constrained by inclusion and exclusion criteria. Each artifact is editable and can be regenerated. Retrieval of titles, abstracts, and metadata starts only after the researcher approves the search strategy (first checkpoint).

\textbf{Phase 2: Screening and extraction.} The researcher inspects the retrieved results, selects the relevant papers, and uploads their PDFs (second checkpoint, inclusion and exclusion). Text is extracted with PyMuPDF, with an OCR fallback for scanned documents, segmented into chunks, embedded with OpenAI's text-embedding-ada-002 model, and stored in a local FAISS index, together with bibliographic metadata. Because only researcher-approved papers enter this pipeline, the evidence base is one the researcher has explicitly curated.

\textbf{Phase 3: Analysis and reporting.} Following practical RAG design principles~\cite{hasan2025engineering}, user queries are embedded and matched against the indexed documents, and the retrieved passages are passed to the LLM as context for grounded answers and synthesis. Based on this evidence and the approved objectives, the \emph{Report Generation Agent} drafts a structured review report with introduction, methodology, analysis, and discussion sections. Researchers can interrogate the report through the question-answering interface and revise individual sections through further prompts. Validating and approving the final report is the third checkpoint, after which it can be exported.

\section{Study Design}
\label{sec:study_design}

This study follows a Design Science Research (DSR) approach, which emphasizes the design, development, and evaluation of artifacts that address practical problems \cite{hevner2004design,peffers2007design}. The objective is to assess the perceived usefulness and usability, and participants’ trust in generated output of the proposed Human--AI collaborative system in supporting SLR workflows while maintaining human oversight.

\subsection{Research Objectives and Questions}

The primary objective of this study is to evaluate how effectively the proposed framework supports researchers in performing SLR tasks and facilitates human--AI collaboration.

To guide the evaluation, the following Research Questions (RQs) are defined:

\begin{tcolorbox}[colback=gray!10,colframe=black,boxrule=0.5pt,arc=2pt]
\textbf{RQ1:} How useful is the proposed Human--AI collaborative multi-agent system in supporting key phases of the SLR process?
\end{tcolorbox}

\noindent\textit{Rationale:} This question examines which SLR activities participants found the system most helpful for, while researchers remain responsible for validation and interpretation.

\begin{tcolorbox}[colback=gray!10,colframe=black,boxrule=0.5pt,arc=2pt]
\textbf{RQ2:} How do participants perceive the system's usability and expected efficiency relative to their usual literature review practice?
\end{tcolorbox}

\noindent\textit{Rationale:} This question assesses perceived usability, workload, and expected time savings; actual task time was not measured.

\begin{tcolorbox}[colback=gray!10,colframe=black,boxrule=0.5pt,arc=2pt]
\textbf{RQ3:} To what extent do participants report trust in AI-generated output without additional human checking, and does this vary across user groups?
\end{tcolorbox}

\noindent\textit{Rationale:} Human checkpoints protect a review only if researchers use them to verify. This question explores how much participants report trusting unchecked output after interacting with a system that provides mandatory checkpoints.

\subsection{Study Design and Procedure}

A user study was conducted in a remote, self-paced setting, where participants interacted with the system independently using a web-based interface. Prior to the study, participants were provided with a short video tutorial explaining the system functionalities and workflow. This was intended to give participants a consistent baseline understanding of how to use the system before performing the assigned tasks.

Participants were then asked to perform a simplified SLR workflow: defining a research topic, generating search queries, retrieving and selecting papers, uploading and analysing documents, using the question-answering interface, and generating a literature review report.

The study duration was approximately 15 minutes per participant. The tasks were designed to simulate realistic usage scenarios while keeping the interaction lightweight and accessible. Participants were allowed to explore the system freely, reflecting natural user behavior. Although the interaction time was limited, the study was designed to capture initial user impressions and evaluate the perceived usability of the system in early-stage literature review tasks.

The participant group primarily consisted of individuals with backgrounds in software engineering, including mostly students and researchers. This matches the intended user population of the system.

\subsection{Data Collection Instruments}

Data were collected through a post-interaction questionnaire that participants completed after finishing the assigned tasks. It combined structured items with open-ended questions and was organized into seven sections: demographic information, usability and interaction, cognitive load and user experience, learning and retention, productivity and workflow, recommendation and concerns, and open feedback. Table~\ref{tab:questionnaire_design} in the appendix summarizes the focus of each section, the number of questions, and the response format.

Most quantitative items used a five-point Likert scale, while selected questions used categorical or binary formats. Together they covered usability, cognitive load, learning support, productivity, perceived retrieval quality, adoption intention, and user concerns. The two open-ended questions asked about system strengths, limitations, and desired improvements, providing qualitative insight into user expectations and perceived gaps in the current system.

\subsection{Data Collection}
A total of 63 participants completed the study, and data collection took place over an approximately eight-week period, from 7 October 2025 to 30 November 2025. Participants were from academic and professional backgrounds in software engineering, including postgraduate students, academic researchers, and industry professionals. The survey was distributed through convenience sampling via the University portal, LinkedIn, and direct email sharing with researchers.

Participation was voluntary and anonymous. Before responding, participants were informed of the study's purpose, the data collected, and their right to withdraw. Data handling followed our institution's research-ethics guidance, collecting only data necessary for the study. This work was conducted for research purposes only and has no commercial agenda; the system is a research prototype rather than a product, and its generative components rely on third-party large language model APIs used under the terms of those services.

Participants could select multiple roles: 40 identified as researchers, 33 as students, 10 as industry professionals, and 1 as a teacher. This participant group was selected because they represent potential users of automated literature review systems. Their experience with academic reading and research workflows allows them to provide meaningful feedback on the usefulness of the proposed system.

\subsection{Data Analysis}
The collected data were analyzed using both quantitative and qualitative methods. Quantitative responses were analyzed using descriptive statistics, including means, medians, variances, standard deviations, and percentage distributions \cite{wohlin2006empirical}.

Qualitative responses from the open-ended survey questions were analyzed using thematic analysis \cite{braun2006using}. The first author conducted the qualitative analysis in Microsoft Excel by first reading all responses for familiarization, then assigning descriptive codes to meaningful response segments, and finally grouping related codes into broader categories through an iterative refinement process. In total, 30 response pairs were reviewed. After excluding blank and non-informative responses, 76 coded segments were identified, resulting in 24 unique codes and 8 broader code groups, which were subsequently refined into five final themes. Table~\ref{tab:code_examples} in the appendix presents illustrative examples from the initial coding stage.


Table~\ref{tab:code_groups} in the appendix shows how related initial codes were organized into broader code groups before being refined into the final themes.

\section{Results}
\label{sec:results}

This section presents the findings of the user study in relation to the three research questions. The results are reported using both quantitative and qualitative evidence from the survey responses.

\subsection{RQ1: Usefulness of the Proposed System}

The survey results indicate that participants perceived the system as particularly useful in early-stage planning and synthesis tasks. The highest mean scores were observed for the intuitiveness of framing or editing research questions ($M=3.90$, $SD=0.80$) and system responsiveness ($M=3.90$, $SD=0.89$), followed by the intuitiveness of framing or editing research objectives ($M=3.81$, $SD=0.76$). These results suggest that users found the system especially helpful when defining the scope of a review and interacting with the system during the planning phase.

Table~\ref{tab:grouped_stats} summarizes the average ratings across the main evaluation dimensions. Usability received the highest grouped mean score (3.78), followed by learning support and productivity (both 3.68). The single item measuring trust in generated outputs without
additional human checking received a mean score of 3.02,
indicating neutral trust on average. This finding is examined
in detail in Section~\ref{sec:disc_rq3}.

\begin{table}[htbp]
\centering
\caption{Grouped descriptive summary of Likert-scale responses (N=63).}
\label{tab:grouped_stats}
\footnotesize
\renewcommand{\arraystretch}{1.15}
\begin{tabularx}{\textwidth}{@{} >{\raggedright\arraybackslash}p{2.6cm} L >{\centering\arraybackslash}p{1.4cm} @{}}
\toprule
\textbf{Dimension} & \textbf{Representative items} & \textbf{Avg. mean} \\
\midrule
Usability & Q\&A interaction, report generation, responsiveness, objective/RQ framing, data extraction & 3.78 \\
\addlinespace
Learning support & Understanding material, key concepts, confidence in applying content & 3.68 \\
\addlinespace
Productivity & Answer accuracy, report summaries, comprehension support & 3.68 \\
\addlinespace
Trust and oversight & Trust in generated outputs without additional human checking & 3.02 \\
\bottomrule
\end{tabularx}

\smallskip
\begin{minipage}{\textwidth}
\scriptsize\raggedright
\textit{Note.} Each value is the mean of the item means in Table~\ref{tab:likert_all}. The focus-on-comprehension item is grouped under Productivity; the negatively worded cognitive-load items and the recommendation items are reported only in Table~\ref{tab:likert_all}.
\end{minipage}
\end{table}


Table~\ref{tab:likert_all} in the appendix reports descriptive statistics for all Likert-scale items. Interaction quality, report generation, and responsiveness were rated positively overall, as were the likelihood of recommending the system ($M = 3.89$) and of continuing to use it ($M = 3.71$). Mental demand was moderate ($M = 3.00$), and participants reported feeling stuck or unsure what to do next at a moderate level ($M = 3.52$). Trust in generated summaries without additional human checking received the lowest rating of the positively worded items ($M = 3.02$).

Overall, the RQ1 findings suggest that the proposed system is perceived as useful across multiple stages of the SLR workflow, especially in planning, interaction, and structured synthesis. Trust in unchecked output was the lowest-rated dimension, which we examine in detail under RQ3.

\subsection{RQ2: Usability and Expected Efficiency}
\label{sec:RQ2}

Participants were asked how much time they expected the system to save compared with a traditional literature review workflow. Nearly half (49.2\%) expected savings of between 25\% and 50\% of the time typically required for literature exploration tasks, 27.0\% expected savings exceeding 50\%, and 23.8\% expected savings below 25\%.

Because the time-savings item offered only positive brackets, we also report two items that allowed neutral or negative responses. Asked whether the system saved time in understanding documents, 66.7\% reported that it significantly reduced time, 28.6\% that it saved a little time, and 4.8\% that it took the same or more time. Compared with their usual methods, 71.4\% found the tool more efficient, 17.5\% about the same, and 11.1\% less efficient.\footnote{Five respondents selected two adjacent options on this item; each is counted under the first option they selected.}

The open-ended responses were examined through thematic analysis, which yielded five themes from 24 unique codes in eight code groups (Table~\ref{tab:thematic_analysis}).

\begin{table}[htbp]
\centering
\caption{Themes identified from open-ended survey responses.}
\label{tab:thematic_analysis}
\footnotesize
\renewcommand{\arraystretch}{1.15}
\begin{tabularx}{\textwidth}{@{} >{\raggedright\arraybackslash}p{2.6cm} L L @{}}
\toprule
\textbf{Theme} & \textbf{Description} & \textbf{Implication} \\
\midrule
Support for research question and search formulation & Participants valued the system's ability to generate research questions, objectives, and search strings. & Indicates perceived usefulness in the planning stage of the SLR workflow. \\
\addlinespace
Report generation and synthesis support & Users appreciated automated synthesis, report generation, and data extraction features. & Suggests the system is especially valued for analysis and structured reporting. \\
\addlinespace
Retrieval quality and coverage & Participants requested broader database integration and more precise search strings. & Highlights retrieval as a key limitation and a priority for future improvement. \\
\addlinespace
Transparency and evidence traceability & Users wanted clearer links between generated outputs and the papers used as evidence. & Reinforces the importance of evidence grounding and explainability in Human--AI systems. \\
\addlinespace
Workflow control and manual checkpoints & Several responses emphasized the need for stronger workflow control and manual validation. & Supports human-in-the-loop design rather than full black-box automation. \\
\bottomrule
\end{tabularx}
\end{table}

Participants valued the system most for planning, synthesis, and report generation, but repeatedly asked for broader retrieval coverage, clearer links between generated outputs and source papers, and stronger manual checkpoints; 69.8\% also said they would prefer more manual intervention options at stages such as screening and synthesis. These requests concern whether users can verify what the system produces, which we examine next through participants' trust in unchecked output.

\subsection{RQ3: Trust in Unchecked Output}
\label{sec:trust}

\begin{figure}[htbp]
\centering
\includegraphics[width=\linewidth]{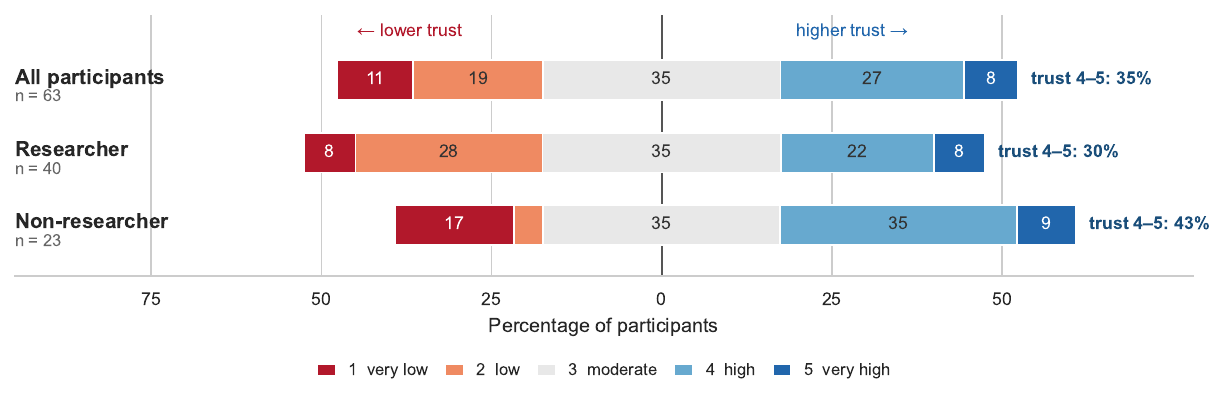}
\caption{Trust in AI-generated output without additional human checking, overall and by Researcher role (Mann--Whitney $U = 400.5$, $p = .38$).}
\label{fig:trust}
\end{figure}

A human checkpoint protects the review only if the researcher uses it to verify rather than to confirm; a checkpoint that is clicked through produces the appearance of oversight without its substance. We therefore examined participants' self-reported trust in AI-generated synthesis without additional checking, using the item \emph{``To what extent do you trust the generated summaries and reports without additional human checking?''} (1 = very low, 5 = very high). Because these data were collected as part of a general usability evaluation, the following analyses are exploratory: we report all three tests we ran with exact $p$-values and avoid causal interpretation.

Figure~\ref{fig:trust} shows the distribution. Responses were spread across the scale (mean 3.02, SD 1.11, median 3) rather than concentrated at low trust. Nineteen of 63 participants (30.2\%) rated their trust in unchecked output as low (1 or 2), 22 (35\%) were neutral, and 22 (35\%) rated it high (4 or 5) after only a short interaction with the system. A high rating reflects a self-reported attitude rather than observed behaviour, and the wording ``without additional human checking'' may have been read as checking beyond the mandatory checkpoints; we therefore do not interpret these ratings as evidence that participants would actually skip verification.

We also examined whether trust in unchecked output was related to expected time savings. The two were positively associated (Spearman $\rho = 0.30$, $p = .017$; Kendall $\tau_b = 0.27$, $p = .015$; $n = 63$). Because no participant completed a full review, the time-savings item cannot reflect a measured reduction in effort. One possible explanation is that both items partly reflect a general attitude toward AI assistance, but our data cannot distinguish this from other explanations; we return to this in Section~\ref{sec:validity}.

Finally, we compared trust between participants who identified as researchers ($n = 40$) and those who did not ($n = 23$); because roles were multi-select and overlapping, we used this single binary split. We could not detect a difference in trust between the two groups (Mann--Whitney $U = 400.5$, $p = .38$; rank-biserial $r = -0.13$, a small effect), and medians were equal (both 3). Descriptively, non-researchers rated 4 or 5 somewhat more often than researchers (43.5\% vs 30.0\%), a descriptive pattern suggesting higher self-reported trust among participants who did not identify as researchers; however, this difference was not statistically detectable. A sensitivity analysis (G*Power~\cite{faul2007gpower}) shows that with groups of 40 and 23, the test had 80\% power ($\alpha = .05$, two-sided) only for effects of about $d \approx 0.8$ or larger. Only large differences would therefore have been reliably detected; small or medium differences, such as the one observed (rank-biserial $r = -0.13$), remain possible.

\section{Discussion}
\label{sec:discussion}

Participants valued CoSLR most for planning, interaction, and synthesis, and rated it positively for usability. At the same time, about a third reported high trust in AI-generated synthesis without additional checking, although the system's value depends on verification taking place. We therefore discuss usefulness and oversight separately.

\subsection{Usefulness and Expected Efficiency (RQ1, RQ2)}

Participants rated research-question and objective framing, responsiveness, and report generation most positively, and the open-ended feedback named the same features. Earlier tools such as Abstrackr and ASReview showed that machine learning can reduce screening effort~\cite{wallace2010semi,van2020asreview}; our participants valued support well beyond screening, in planning, search formulation, document interaction, and reporting. This is consistent with reviews reporting that SLR tools remain fragmented across stages~\cite{khalil2022tools,van2021automation}, and suggests that part of a system's usefulness comes from connecting stages rather than automating one.

Participants also expected the system to save time: most expected savings of at least 25\%. On the two items that allowed neutral or negative answers, 4.8\% and 28.6\% of participants respectively reported no gain or a loss. Because no participant completed a full review, these are expectations rather than measured savings; establishing actual efficiency requires a task-matched comparison.

The feedback also qualified this picture. Participants asked for broader retrieval coverage, explicit links between generated text and source papers, and more control over intermediate outputs, consistent with guidelines that users should be able to inspect and correct AI output~\cite{amershi2019guidelines}. Retrieval-augmented generation grounds output internally, but users need that grounding made visible before they can judge whether to rely on it.

\subsection{Trust in Unchecked Output (RQ3)}
\label{sec:disc_rq3}

Trust in AI-generated summaries without additional checking had the lowest mean of the positively worded items (3.02), but it was not uniformly low: responses spread across the scale, and 22 of 63 participants (34.9\%) rated it 4 or 5 after only a short interaction. On average participants were neutral, but a substantial minority reported high trust despite the mandatory checkpoints, so oversight cannot be assumed simply because an interface provides the opportunity for it.

This high trust coexisted with positive usability ratings. The properties that make an assistant pleasant to use---fluency, responsiveness, and confident, well-structured output---may also reduce a user's felt need to check. Bu\c{c}inca et al.~\cite{buccinca2021trust} found that cognitive forcing interventions reduce over-reliance but are rated less favourably than interfaces that let users act quickly. A single confirmation step demands far less engagement than the interventions they tested, and we did not compare checkpoint designs, so our results are consistent with this tension rather than evidence for it. They do show that a usability score alone is not evidence that oversight is well calibrated.

Trust was also positively associated with expected time savings ($\rho = 0.30$). One possible explanation is a general disposition toward AI assistance underlying both, but our data cannot establish this. Finally, non-researchers reported high trust somewhat more often than researchers (43.5\% vs 30.0\%), but the difference was not statistically detectable, and we treat it only as a hypothesis for an adequately powered study.

\subsection{Implications for Human--AI SLR Systems}

\paragraph{Checkpoint strength as a design variable.} CoSLR applies the same checkpoints to every stage, and each can be satisfied by a single confirmation. If a substantial share of users report high trust in unchecked output, such a checkpoint may certify that a step was reached without ensuring that verification occurred. The engagement a checkpoint demands should therefore be a design variable: more effortful checks, such as opening a cited source or confirming specific extracted claims against the evidence, could be reserved for the stages where an unverified error is most consequential, such as inclusion decisions and final synthesis. Because we found no statistically detectable difference by researcher role, stage risk is a more defensible basis for checkpoint strength than user role.

\paragraph{Visible provenance and retrieval quality.} Evidence grounding should be exposed in the interface and in generated reports, not only used to condition generation, so that verification at each checkpoint is practical. Because missed studies propagate into every later stage, retrieval coverage also bounds the trustworthiness of the synthesis built on it.

\paragraph{Licensing.} CoSLR sends text from uploaded papers to cloud-hosted model services, so whether this is permitted depends on each document's licence and the service's terms. The prototype keeps the vector index local (FAISS) but does not check licences, leaving compliance with the user; future versions should warn users before external processing and support locally hosted models where licensing requires it.

\section{Threats to Validity}
\label{sec:validity}

We discuss construct, internal, external, and conclusion validity following Runeson and H\"ost~\cite{runeson2009guidelines}.

\paragraph{Construct validity.} The evaluation relied on post-interaction self-report rather than measured task performance or observed behaviour. Trust was measured with a single item whose wording (``without additional human checking'') may have been read as checking beyond the mandatory checkpoints, so high ratings cannot be equated with skipping verification. The time-savings item offered only positive brackets and no participant completed a full review, so it captures expected rather than measured savings; two further items that allowed neutral or negative answers showed a similar but less uniformly positive pattern. Finally, the quality of the system's own outputs, such as retrieval recall and citation faithfulness, was not evaluated, so we cannot relate participants' trust to the actual reliability of the outputs they saw. The qualitative coding was performed by a single author without a second coder, so the themes may reflect one coder's interpretation.

\paragraph{Internal validity.} The study was remote, self-paced, and short (about 15 minutes), which limited control over how carefully participants followed the tutorial and tasks, and may not reflect use during a complete review. There was no comparison condition with traditional or commercial tools, and participants were not required to have SLR experience, so for some the session captured first impressions rather than an informed comparison with their usual practice.

\paragraph{External validity.} Participants came mainly from software engineering, so the findings may not generalise to fields such as medicine, social sciences, or law, where review practices differ. The prototype integrates a limited set of data sources, and the results describe its current maturity rather than that of a mature SLR platform.

\paragraph{Conclusion validity.} The trust analyses re-use data collected for a usability evaluation and were not preregistered; they are exploratory, and we report all tests we ran with exact $p$-values. With $N = 63$, the subgroup test could reliably detect only large effects ($d \approx 0.8$ at 80\% power). All relationships are correlational and were analysed with non-parametric methods suited to ordinal data, so we draw no causal conclusions.
\section{Conclusion}
\label{sec:conclusion}
This paper presented CoSLR, a Human--AI multi-agent system that supports planning, retrieval, document interaction, synthesis, and report generation in one workflow, with mandatory human checkpoints between AI-generated output and its acceptance. In an exploratory study with 63 participants, the system was rated positively for usability and for research-question formulation, search support, responsiveness, and report generation. Yet 35\% of participants reported high trust in AI-generated summaries and reports without additional checking after only a short interaction, and this trust was associated with higher expected time savings. High usability is therefore not, on its own, evidence of well-calibrated oversight.

Three directions follow. First, checkpoint strength should be treated as a design variable, with more effortful verification at the stages where unverified errors are most consequential. Second, evidence provenance should be visible in the interface and in generated reports, so that verification is practical rather than nominal. Third, future evaluations should measure output reliability alongside users' trust, compare checkpoint designs directly, and observe verification behaviour during complete reviews rather than relying on self-report. Human--AI systems can make systematic reviews more manageable, but their value depends on whether the oversight they rely on actually takes place---a problem that system design must address directly rather than assume.
\section*{Data and Code Availability}

The replication package contains the anonymised survey responses, the full questionnaire, the agent prompts and system configuration, and an analysis notebook that reproduces the reported statistics, tables, and figures. The package is publicly available at \cite{slrv2github}.
\section*{Declaration on Generative AI}
Generative AI tools were used in preparing this manuscript in the following ways: Claude (Anthropic) and ChatGPT (OpenAI) were used to improve the wording and clarity of the text, to help format tables and LaTeX code. All AI-assisted text and analyses were reviewed, verified and revised by the authors, who take full responsibility for the content of this paper. No generative AI tool was used to produce the study data, the participant responses, or the qualitative coding.
%
\bibliographystyle{splncs04}
\bibliography{references}

@techreport{kitchenham2007guidelines,
  title       = {Guidelines for performing systematic literature reviews in software engineering},
  author      = {Kitchenham, Barbara and Charters, Stuart},
  year        = {2007},
  number      = {EBSE-2007-01},
  institution = {Keele University and Durham University},
}

@article{page2021prisma,
  title={The PRISMA 2020 statement: an updated guideline for reporting systematic reviews},
  author={Page, Matthew J and McKenzie, Joanne E and Bossuyt, Patrick M and Boutron, Isabelle and Hoffmann, Tammy C and Mulrow, Cynthia D and Shamseer, Larissa and Tetzlaff, Jennifer M and Akl, Elie A and Brennan, Sue E and others},
  journal={bmj},
  volume={372},
  year={2021},
  publisher={British Medical Journal Publishing Group}
}

@article{bommasani2021opportunities,
  title={On the opportunities and risks of foundation models},
  author={Bommasani, Rishi and Hudson, Drew A and Adeli, Ehsan and Altman, Russ and Arora, Simran and von Arx, Sydney and Bernstein, Michael S and Bohg, Jeannette and Bosselut, Antoine and Brunskill, Emma and others},
  journal={arXiv preprint arXiv:2108.07258},
  year={2021}
}

@article{okamura2020adaptive,
  title={Adaptive trust calibration for human-AI collaboration},
  author={Okamura, Kazuo and Yamada, Seiji},
  journal={Plos one},
  volume={15},
  number={2},
  pages={e0229132},
  year={2020},
  publisher={Public Library of Science San Francisco, CA USA}
}

@inproceedings{ueno2022trust,
  title={Trust in human-AI interaction: Scoping out models, measures, and methods},
  author={Ueno, Takane and Sawa, Yuto and Kim, Yeongdae and Urakami, Jacqueline and Oura, Hiroki and Seaborn, Katie},
  booktitle={CHI conference on human factors in computing systems extended abstracts},
  pages={1--7},
  year={2022}
}

@article{brown2020language,
  title={Language models are few-shot learners},
  author={Brown, Tom and Mann, Benjamin and Ryder, Nick and Subbiah, Melanie and Kaplan, Jared D and Dhariwal, Prafulla and Neelakantan, Arvind and Shyam, Pranav and Sastry, Girish and Askell, Amanda and others},
  journal={Advances in neural information processing systems},
  volume={33},
  pages={1877--1901},
  year={2020}
}

@article{van2020asreview,
  title   = {Open source software for efficient and transparent reviews},
  author  = {van de Schoot, Rens and de Bruin, Jonathan and Schram, Raoul and Zahedi, Parisa and de Boer, Jan and Weijdema, Felix and Kramer, Bianca and Huijts, Martijn and Hoogerwerf, Maarten and Ferdinands, Gerbrich and others},
  journal = {Nature Machine Intelligence},
  volume  = {3},
  number  = {2},
  pages   = {125--133},
  year    = {2021},
  publisher = {Nature Publishing Group},
}

@inproceedings{wallace2012abstrackr,
author = {Wallace, Byron C. and Small, Kevin and Brodley, Carla E. and Lau, Joseph and Trikalinos, Thomas A.},
title = {Deploying an interactive machine learning system in an evidence-based practice center: abstrackr},
year = {2012},
isbn = {9781450307819},
publisher = {Association for Computing Machinery},
address = {New York, NY, USA},
doi = {10.1145/2110363.2110464},
booktitle = {Proceedings of the 2nd ACM SIGHIT International Health Informatics Symposium},
pages = {819–824},
numpages = {6},
location = {Miami, Florida, USA},
series = {IHI '12}
}

@article{lewis2020retrieval,
  title={Retrieval-augmented generation for knowledge-intensive nlp tasks},
  author={Lewis, Patrick and Perez, Ethan and Piktus, Aleksandra and Petroni, Fabio and Karpukhin, Vladimir and Goyal, Naman and K{\"u}ttler, Heinrich and Lewis, Mike and Yih, Wen-tau and Rockt{\"a}schel, Tim and others},
  journal={Advances in neural information processing systems},
  volume={33},
  pages={9459--9474},
  year={2020}
}

@article{ouzzani2016rayyan,
  title={Rayyan—a web and mobile app for systematic reviews},
  author={Ouzzani, Mourad and Hammady, Hossam and Fedorowicz, Zbys and Elmagarmid, Ahmed},
  journal={Systematic reviews},
  volume={5},
  number={1},
  pages={210},
  year={2016},
  publisher={Springer}
}

@inproceedings{barnett2024seven,
  title={Seven failure points when engineering a retrieval augmented generation system},
  author={Barnett, Scott and Kurniawan, Stefanus and Thudumu, Srikanth and Brannelly, Zach and Abdelrazek, Mohamed},
  booktitle={Proceedings of the IEEE/ACM 3rd International Conference on AI Engineering-Software Engineering for AI},
  pages={194--199},
  year={2024}
}

@inproceedings{amershi2019guidelines,
  title={Guidelines for human-AI interaction},
  author={Amershi, Saleema and Weld, Dan and Vorvoreanu, Mihaela and Fourney, Adam and Nushi, Besmira and Collisson, Penny and Suh, Jina and Iqbal, Shamsi and Bennett, Paul N and Inkpen, Kori and others},
  booktitle={Proceedings of the 2019 chi conference on human factors in computing systems},
  pages={1--13},
  year={2019}
}

@article{buccinca2021trust,
  title={To trust or to think: cognitive forcing functions can reduce overreliance on AI in AI-assisted decision-making},
  author={Bu{\c{c}}inca, Zana and Malaya, Maja Barbara and Gajos, Krzysztof Z},
  journal={Proceedings of the ACM on Human-computer Interaction},
  volume={5},
  number={CSCW1},
  pages={1--21},
  year={2021},
  publisher={ACM New York, NY, USA}
}

@inproceedings{bansal2019updates,
  title={Updates in human-ai teams: Understanding and addressing the performance/compatibility tradeoff},
  author={Bansal, Gagan and Nushi, Besmira and Kamar, Ece and Weld, Daniel S and Lasecki, Walter S and Horvitz, Eric},
  booktitle={Proceedings of the AAAI conference on artificial intelligence},
  volume={33},
  pages={2429--2437},
  year={2019}
}

@inproceedings{hasan2025engineering,
  title={Engineering rag systems for real-world applications: Design, development, and evaluation},
  author={Hasan, Md Toufique and Waseem, Muhammad and Kemell, Kai-Kristian and Khan, Ayman Asad and Saari, Mika and Abrahamsson, Pekka},
  booktitle={Euromicro Conference on Software Engineering and Advanced Applications},
  pages={143--158},
  year={2025},
  organization={Springer}
}

@article{khalil2022tools,
  title={Tools to support the automation of systematic reviews: a scoping review},
  author={Khalil, Hanan and Ameen, Daniel and Zarnegar, Armita},
  journal={Journal of clinical epidemiology},
  volume={144},
  pages={22--42},
  year={2022},
  publisher={Elsevier}
}

@article{van2021automation,
  title={Automation of systematic literature reviews: A systematic literature review},
  author={Van Dinter, Raymon and Tekinerdogan, Bedir and Catal, Cagatay},
  journal={Information and software technology},
  volume={136},
  pages={106589},
  year={2021},
  publisher={Elsevier}
}

@article{wallace2010semi,
  title={Semi-automated screening of biomedical citations for systematic reviews},
  author={Wallace, Byron C and Trikalinos, Thomas A and Lau, Joseph and Brodley, Carla and Schmid, Christopher H},
  journal={BMC bioinformatics},
  volume={11},
  number={1},
  pages={55},
  year={2010},
  publisher={Springer}
}

@article{hamel2020evaluation,
  title={An evaluation of DistillerSR’s machine learning-based prioritization tool for title/abstract screening--impact on reviewer-relevant outcomes},
  author={Hamel, C and Kelly, SE and Thavorn, K and Rice, DB and Wells, GA and Hutton, B},
  journal={BMC medical research methodology},
  volume={20},
  number={1},
  pages={256},
  year={2020},
  publisher={Springer}
}

@article{marshall2016robotreviewer,
  title={RobotReviewer: evaluation of a system for automatically assessing bias in clinical trials},
  author={Marshall, Iain J and Kuiper, Jo{\"e}l and Wallace, Byron C},
  journal={Journal of the American Medical Informatics Association},
  volume={23},
  number={1},
  pages={193--201},
  year={2016},
  publisher={Oxford University Press}
}

@inproceedings{wang2024zero,
  title={Zero-shot generative large language models for systematic review screening automation},
  author={Wang, Shuai and Scells, Harrisen and Zhuang, Shengyao and Potthast, Martin and Koopman, Bevan and Zuccon, Guido},
  booktitle={European Conference on Information Retrieval},
  pages={403--420},
  year={2024},
  organization={Springer}
}

@inproceedings{huotala2024promise,
  title={The promise and challenges of using LLMs to accelerate the screening process of systematic reviews},
  author={Huotala, Aleksi and Kuutila, Miikka and Ralph, Paul and M{\"a}ntyl{\"a}, Mika},
  booktitle={Proceedings of the 28th International Conference on Evaluation and Assessment in Software Engineering},
  pages={262--271},
  year={2024}
}

@article{sami2024system,
  title={System for systematic literature review using multiple ai agents: Concept and an empirical evaluation},
  author={Sami, Abdul Malik and Rasheed, Zeeshan and Kemell, Kai-Kristian and Waseem, Muhammad and Kilamo, Terhi and Saari, Mika and Duc, Anh Nguyen and Syst{\"a}, Kari and Abrahamsson, Pekka},
  journal={arXiv preprint arXiv:2403.08399},
  year={2024}
}

@article{rouzrokh2025lattereview,
  title={LatteReview: a multi-agent framework for systematic review automation using large language models},
  author={Rouzrokh, Pouria and Khosravi, Bardia and Rouzrokh, Parsa and Shariatnia, Moein},
  journal={arXiv preprint arXiv:2501.05468},
  year={2025}
}

@article{hevner2004design,
  title={Design science in information systems research1},
  author={Hevner, Alan R and March, Salvatore T and Park, Jinsoo and Ram, Sudha},
  journal={MIS quarterly},
  volume={28},
  number={1},
  pages={75--106},
  year={2004},
  publisher={Management Information Systems Research Center, University of Minnesota}
}

@article{peffers2007design,
  title={A design science research methodology for information systems research},
  author={Peffers, Ken and Tuunanen, Tuure and Rothenberger, Marcus A and Chatterjee, Samir},
  journal={Journal of management information systems},
  volume={24},
  number={3},
  pages={45--77},
  year={2007},
  publisher={Taylor \& Francis}
}

@inproceedings{han2024rag,
  title={Rag-qa arena: Evaluating domain robustness for long-form retrieval augmented question answering},
  author={Han, Rujun and Zhang, Yuhao and Qi, Peng and Xu, Yumo and Wang, Jenyuan and Liu, Lan and Wang, William Yang and Min, Bonan and Castelli, Vittorio},
  booktitle={Proceedings of the 2024 conference on empirical methods in natural language processing},
  pages={4354--4374},
  year={2024}
}

@manual{eppireviewer2025,
  author       = {James Thomas and Sergio Graziosi and James Brunton and Zhe Ghouze and Paul O'Driscoll and Melissa Bond and Anna Koryakina},
  title        = {{EPPI-Reviewer}: Advanced Software for Systematic Reviews, Maps and Evidence Synthesis},
  organization = {EPPI-Centre, UCL Social Research Institute, University College London},
  year         = {2023}
}

@article{song2026paperorchestra,
  title={Paperorchestra: A multi-agent framework for automated ai research paper writing},
  author={Song, Yiwen and Song, Yale and Pfister, Tomas and Yoon, Jinsung},
  journal={arXiv preprint arXiv:2604.05018},
  year={2026}
}

@incollection{wohlin2006empirical,
  title={Empirical research methods in web and software engineering},
  author={Wohlin, Claes and H{\"o}st, Martin and Henningsson, Kennet},
  booktitle={Web engineering},
  pages={409--430},
  year={2006},
  publisher={Springer}
}

@article{braun2006using,
  title={Using thematic analysis in psychology},
  author={Braun, Virginia and Clarke, Victoria},
  journal={Qualitative research in psychology},
  volume={3},
  number={2},
  pages={77--101},
  year={2006},
  publisher={Taylor \& Francis}
}

@misc{slrv2github,
  author       = {{GPT Laboratory}},
  title        = {{slr-v2}},
  howpublished = {GitHub repository},
  year         = {2026},
  url          = {https://github.com/GPT-Laboratory/slr-v2},
  note         = {Accessed: 2026-06-08}
}

@article{runeson2009guidelines,
  title={Guidelines for conducting and reporting case study research in software engineering},
  author={Runeson, Per and H{\"o}st, Martin},
  journal={Empirical software engineering},
  volume={14},
  number={2},
  pages={131--164},
  year={2009},
  publisher={Springer}
}

@article{faul2007gpower,
  title={G*Power 3: A flexible statistical power analysis program for the social, behavioral, and biomedical sciences},
  author={Faul, Franz and Erdfelder, Edgar and Lang, Albert-Georg and Buchner, Axel},
  journal={Behavior Research Methods},
  volume={39},
  number={2},
  pages={175--191},
  year={2007},
  publisher={Springer},
  doi={10.3758/BF03193146}
}

@article{joos2026leveraging,
     author  = {Joos, Lucas and Keim, Daniel A. and Fischer, Maximilian T.},
     title   = {Leveraging {LLMs} for semi-automatic corpus filtration in systematic literature reviews},
     journal = {Computers \& Graphics},
     volume={135},
     year    = {2026},
}

@article{ye2024hybrid,
     author  = {Ye, Anjia and Maiti, Ananda and Schmidt, Matthew and Pedersen, Scott J.},
     title   = {A Hybrid Semi-Automated Workflow for Systematic and Literature Review Processes with Large Language Model Analysis},
     journal = {Future Internet},
     volume  = {16},
     number  = {5},
     pages   = {167},
     year    = {2024},
}

\appendix
\setcounter{figure}{0}\renewcommand{\thefigure}{A\arabic{figure}}
\setcounter{table}{0}\renewcommand{\thetable}{A\arabic{table}}
\section{Additional Figures and Tables}

\begin{figure}[htbp]
\centering
\includegraphics[width=\textwidth]{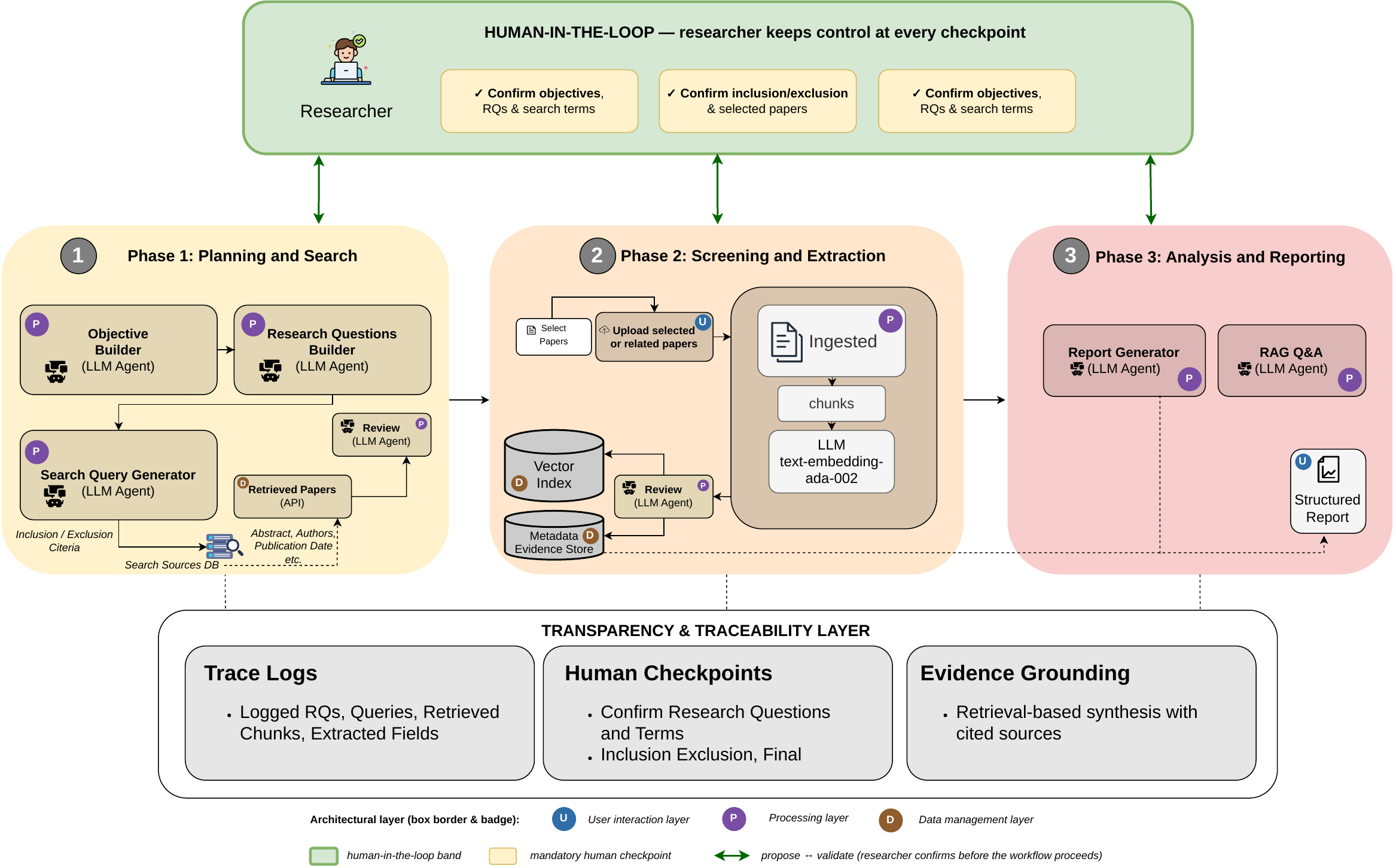}
\caption{High-level architecture of CoSLR: three phases (planning and search, screening and extraction, analysis and reporting) with human checkpoints and trace logs.}
\label{fig:architecture}
\end{figure}

\begin{table}[htbp]
\centering
\caption{Structure of the post-interaction questionnaire.}
\label{tab:questionnaire_design}
\footnotesize
\renewcommand{\arraystretch}{1.15}
\begin{tabularx}{\textwidth}{@{}
  >{\raggedright\arraybackslash}p{2.4cm}
  L
  >{\centering\arraybackslash}p{0.9cm}
  >{\raggedright\arraybackslash}p{2.2cm} @{}}
\toprule
\textbf{Section} & \textbf{Focus / Measured Dimension} & \textbf{Items} & \textbf{Question Type} \\
\midrule
Demographic information & Participant role, document usage frequency, explored content, and usage setting. & 4 & Categorical / multiple choice \\
\addlinespace
Usability and interaction & Support for defining research objectives and questions, generating search strings, screening articles, report generation, and question-answering interaction. & 10 & Likert / binary \\
\addlinespace
Cognitive load and user experience & Mental demand, uncertainty during use, and whether the system reduces technical handling effort. & 3 & Likert \\
\addlinespace
Learning and retention & Perceived support for understanding material, identifying key concepts, and confidence in discussing or applying content. & 3 & Likert \\
\addlinespace
Productivity and workflow & Perceived accuracy, usefulness of summaries, efficiency, time savings, retrieval breadth, risk of missing studies, and preferred level of manual intervention. & 8 & Likert / categorical \\
\addlinespace
Recommendation and concerns & Adoption intention, preference over traditional approaches, and privacy, security, or ethical concerns. & 3 & Likert / binary \\
\addlinespace
Open feedback & Strengths, limitations, and desired improvements. & 2 & Open-ended \\
\midrule
\textbf{Total} & & \textbf{33} & Mixed format \\
\bottomrule
\end{tabularx}
\end{table}

\begin{table}[htbp]
\centering
\caption{Illustrative examples of initial coding in the qualitative analysis.}
\label{tab:code_examples}
\footnotesize
\renewcommand{\arraystretch}{1.15}
\begin{tabularx}{\textwidth}{@{} L >{\raggedright\arraybackslash}p{3.6cm} @{}}
\toprule
\textbf{Response Excerpt} & \textbf{Initial Code} \\
\midrule
``Add more data sources like IEEE'' & More data sources \\
``Search string need to be more precise'' & Query precision \\
``The created report properly cited the papers'' & Citation traceability \\
``Would like a few more and clearer checkpoints'' & Manual checkpoints \\
\bottomrule
\end{tabularx}
\end{table}

\begin{table}[htbp]
\centering
\caption{Illustrative code groups used in the qualitative analysis.}
\label{tab:code_groups}
\footnotesize
\renewcommand{\arraystretch}{1.15}
\begin{tabularx}{\textwidth}{@{}
  >{\raggedright\arraybackslash}p{2.4cm}
  L
  >{\raggedright\arraybackslash}p{3.6cm} @{}}
\toprule
\textbf{Code Group} & \textbf{Example Codes} & \textbf{Related Theme} \\
\midrule
Planning support & RQ generation, objective formulation, search query generation & Support for research question and search formulation \\
\addlinespace
Synthesis support & Report generation, data extraction, structured output, summaries & Report generation and synthesis support \\
\addlinespace
Retrieval limitations & More data sources, query precision, missing papers, direct PDF access & Retrieval quality and coverage \\
\addlinespace
Evidence transparency & Citation traceability, source visibility, seeing analyzed papers & Transparency and evidence traceability \\
\addlinespace
User control & Manual checkpoints, workflow separation, validation control, context access issues & Workflow control and manual checkpoints \\
\bottomrule
\end{tabularx}
\end{table}

\begin{table}[htbp]
\centering
\caption{Descriptive statistics for all five-point Likert items (N=63).}
\label{tab:likert_all}
\footnotesize
\renewcommand{\arraystretch}{1.1}
\begin{tabularx}{\textwidth}{@{} L rrrr @{}}
\toprule
\textbf{Item} & \textbf{Mean} & \textbf{Var.} & \textbf{SD} & \textbf{Median} \\
\midrule
\multicolumn{5}{@{}l}{\textit{Usability and interaction}} \\
Intuitiveness of framing/editing research objectives & 3.81 & 0.58 & 0.76 & 4 \\
Intuitiveness of framing/editing research questions & 3.90 & 0.64 & 0.80 & 4 \\
Ease of asking questions and receiving useful answers & 3.62 & 0.79 & 0.89 & 4 \\
Ease of extracting data from included studies & 3.70 & 0.76 & 0.87 & 4 \\
Intuitiveness of report generation & 3.71 & 0.66 & 0.81 & 4 \\
Responsiveness / speed & 3.90 & 0.80 & 0.89 & 4 \\
\addlinespace
\multicolumn{5}{@{}l}{\textit{Cognitive load and user experience}} \\
Mental demand$^\dagger$ & 3.00 & 1.19 & 1.09 & 3 \\
Frequency of feeling stuck or unsure what to do next$^\dagger$ & 3.52 & 0.90 & 0.95 & 3 \\
Focus on comprehension rather than technical handling & 3.67 & 0.87 & 0.93 & 4 \\
\addlinespace
\multicolumn{5}{@{}l}{\textit{Learning and retention}} \\
Improved understanding of the material & 3.78 & 0.69 & 0.83 & 4 \\
Identifying key concepts more clearly than manual reading & 3.60 & 0.89 & 0.94 & 4 \\
Confidence discussing or applying the material & 3.67 & 1.00 & 1.00 & 4 \\
\addlinespace
\multicolumn{5}{@{}l}{\textit{Productivity and workflow}} \\
Relevance / accuracy of answers & 3.57 & 0.80 & 0.89 & 4 \\
Usefulness of report summaries & 3.79 & 0.75 & 0.86 & 4 \\
Trust in summaries and reports without additional human checking & 3.02 & 1.24 & 1.11 & 3 \\
\addlinespace
\multicolumn{5}{@{}l}{\textit{Recommendation and concerns}} \\
Likelihood of recommending to colleagues & 3.89 & 1.00 & 1.00 & 4 \\
Likelihood of continued use & 3.71 & 1.01 & 1.01 & 4 \\
\bottomrule
\end{tabularx}

\smallskip
\begin{minipage}{\textwidth}
\scriptsize\raggedright
\textit{Note.} 1 = lowest, 5 = highest. $^\dagger$Higher values indicate a less favourable experience.
\end{minipage}
\end{table}

\end{document}